\def\year{2019}\relax
\documentclass[letterpaper]{article} %
\usepackage{aaai19}  %
\usepackage{times}  %
\usepackage{helvet}  %
\usepackage{courier}  %
\usepackage{url}  %
\usepackage{graphicx}  %

\usepackage{bm}
\usepackage{enumitem}
\usepackage{xspace}
\usepackage{booktabs}
\usepackage{amsmath} %
\usepackage{amssymb} %
\usepackage{amsfonts}
\usepackage{multirow}
\usepackage{subcaption}
\usepackage[linesnumbered,ruled,vlined]{algorithm2e}

\SetCommentSty{mycommfont}

\newcommand{\tp}[1]{#1^\top} %
\newcommand{\significant}{$ {}^{\text{\textbf{*}}} $}

\newcommand{\unlabeled}{$ {}^{\dag} $}
\newcommand{\citet}[1]{\citeauthor{#1}~\shortcite{#1}}

\newcommand{\model}{\textsc{MEIN}}
\newcommand{\mein}{\textsc{MEIN}}
\newcommand{\imitator}{\textsc{IMN}}
\newcommand{\expert}{\textsc{EXN}}
\newcommand{\strongbase}{\textsc{LM-LSTM}}
\newcommand{\weakbase}{\textsc{LSTM}}
\newcommand{\adv}{\textsc{ADV-LM-LSTM}}
\newcommand{\vat}{\textsc{VAT-LM-LSTM}}
\newcommand{\elmo}{\textsc{ELMo-LSTM}}
\newcommand{\ivat}{i\textsc{VAT-LSTM}}

\DeclareMathOperator*{\LSTMU}{\mathrm{LSTM}}

\DeclareMathOperator*{\ReLU}{\mathrm{ReLU}}
\DeclareMathOperator*{\argmin}{arg\,min}

\nocopyright

\begin{document}
\title{Mixture of Expert/Imitator Networks:\\Scalable Semi-supervised Learning Framework}
\author{
Shun Kiyono$^{\,1}$ \and Jun Suzuki$^{\,1,2}$ \and Kentaro Inui$^{\,1,2}$ \\
  ${}^{1}$ Tohoku University ~~ ${}^{2}$ RIKEN Center for Advanced Intelligence Project \\
 \texttt{\{kiyono,jun.suzuki,inui\}@ecei.tohoku.ac.jp}
}
\maketitle
\begin{abstract}
The current success of deep neural networks (DNNs) in an increasingly broad range of tasks involving artificial intelligence strongly depends on the quality and quantity of labeled training data.
In general, the scarcity of labeled data, which is often observed in many natural language processing tasks, is one of the most important issues to be addressed.
Semi-supervised learning (SSL) is a promising approach to overcoming this issue by incorporating a large amount of unlabeled data.
In this paper, we propose a novel scalable method of SSL for text classification tasks.
The unique property of our method, Mixture of Expert/Imitator Networks, is that imitator networks learn to ``imitate'' the estimated label distribution of the expert network over the unlabeled data, which potentially contributes a set of features for the classification.
Our experiments demonstrate that the proposed method consistently improves the performance of several types of baseline DNNs.
We also demonstrate that our method has the \textit{more data, better performance} property with promising scalability to the amount of unlabeled data.
\end{abstract}

\section{Introduction}
It is commonly acknowledged that deep neural networks (DNNs) can achieve excellent performance in many tasks across numerous research fields, such as image classification~\cite{he:2016:IEEE}, speech recognition~\cite{amodei:2016:ICML}, and machine translation~\cite{wu:2016:arxiv}.
Recent progress in these tasks has been primarily driven by the following two factors:
(1) A large amount of labeled training data exists.
For example, ImageNet~\cite{deng:2009:imagenet}, one of the major datasets for image classification, consists of approximately 14 million labeled images.
(2) DNNs have the property of achieving better performance when trained on a larger amount of labeled training data, namely, the \textit{more data, better performance} property.

However, collecting a sufficient amount of labeled training data is not always easy for many actual applications.
We refer to this issue as the \textit{labeled data scarcity} issue.
This issue is particularly crucial in the field of natural language processing (NLP), where only a few thousand or even a few hundred labeled data are available for most tasks.
This is because, in typical NLP tasks, creating the labeled data often requires the professional supervision of several highly skilled annotators.
As a result, the cost of data creation is high relative to the amount of data.

\begin{figure}[t!]
    \center
    \includegraphics[width=\hsize]{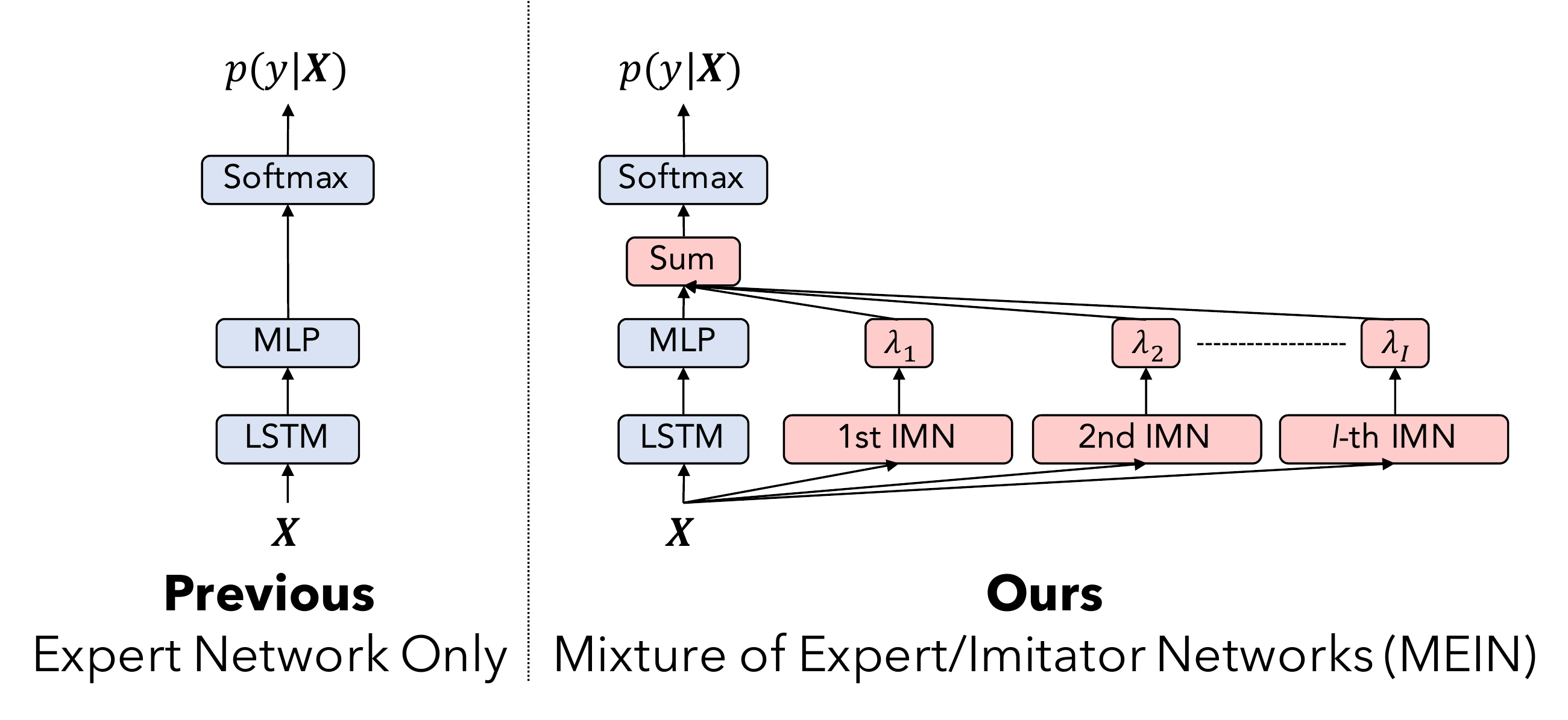}
    \caption{
    Overview of our framework: the Mixture of Expert/Imitator Networks (\model{})
    }
    \label{fig:framework}
\end{figure}
Unlike labeled data, unlabeled data for NLP tasks is essentially a collection of raw texts; thus, an enormous amount of unlabeled data can be obtained from the Internet, such as through the Common Crawl website\footnote{\url{http://commoncrawl.org}}, at a relatively low cost.
With this background, semi-supervised learning (SSL), which leverages unlabeled data in addition to labeled training data for training the parameters of DNNs, is one of the promising approaches to practically addressing the labeled data scarcity issue in NLP.
In fact, some intensive studies have recently been undertaken with the aim of developing SSL methods for DNNs and have shown promising results~\cite{mikolov:2013:NIPS,dai:2015:NIPS,miyato:2017:ICLR,clark:2018:ICLR,peters:2018:NAACL}.

In this paper, we also follow this line of research topic, i.e., discussing SSL suitable for NLP.
Our interest lies in the \textit{more data, better performance} property of the SSL approach over the unlabeled data, which has been implicitly demonstrated in several previous studies~\cite{pennington:2014:EMNLP,peters:2018:NAACL}.
In order to take advantage of the huge amount of unlabeled data and improve performance, we need an SSL approach that scales with the amount of unlabeled data.
However, the scalability of an SSL approach has not yet been widely discussed, since the primary focus of many of the recent studies on SSL in DNNs has been on improving the performance.
For example, several studies have utilized unlabeled data as additional training data, which essentially increases the computational cost of (often complex) DNNs~\cite{miyato:2017:ICLR,clark:2018:ICLR,sato:2018:IJCAI}.
Another SSL approach is to (pre-)train a gigantic bidirectional language model~\cite{peters:2018:NAACL}.
Nevertheless, it has been reported that the training of such a network requires 3 weeks using 32 GPUs~\cite{jozefowicz:2016:arxiv}.
By developing a scalable SSL method, we hope to broaden the usefulness and applicability of DNNs since, as mentioned above, the amount of unlabeled data can be easily increased.

In this paper, we propose a novel scalable method of SSL, which we refer to as the Mixture of Expert/Imitator Networks (\model{}).
Figure~\ref{fig:framework} gives an overview of the \model{} framework, which consists of an expert network (\expert{}) and at least one imitator network (\imitator{}).
To ensure scalability, we design each \imitator{} to be computationally simpler than the \expert{}.
Moreover, we use unlabeled data exclusively for training each \imitator{}; we train the \imitator{} so that it \textit{imitates} the label estimation of the \expert{} over the unlabeled data.
The basic idea underlying the \imitator{} is that we force it to perform the imitation with only a limited view of the given input.
In this way, the \imitator{} effectively learns a set of features, which potentially contributes to the \expert{}.
Intuitively, our method can be interpreted as a variant of several training techniques of DNNs, such as the mixture-of-experts~\cite{jacobs:1991:neural,shazeer:2017:ICLR}, knowledge distillation~\cite{ba:2014:NIPS,hinton:2014:NIPS}, and ensemble techniques.

We conduct experiments on well-studied text classification datasets to evaluate the effectiveness of the proposed method.
We demonstrate that the \model{} framework consistently improves the performance for three distinct settings of the \expert{}.
We also demonstrate that our method has the \textit{more data, better performance} property with promising scalability to the amount of unlabeled data.
In addition, a current popular SSL approach in NLP is to pre-train the language model and then apply it to downstream tasks~\cite{mikolov:2013:NIPS,dai:2015:NIPS,mccann:2017:NIPS,peters:2017:ACL,peters:2018:NAACL}.
We empirically prove in our experiments that \model{} can be easily combined with this approach to further improve the performance of DNNs.

\section{Related Work}
There have been several previous studies in which SSL has been applied to text classification tasks.
A common approach is to utilize unlabeled data as additional training data of the DNN.
Studies employing this approach mainly focused on developing a means of effectively acquiring a teaching signal from the unlabeled data.
For example, in virtual adversarial training (VAT)~\cite{miyato:2017:ICLR} the perturbation is computed from unlabeled data to make the baseline DNN more robust against noise.
\citet{sato:2018:IJCAI} proposed an extension of VAT that generates a more interpretable perturbation.
In addition, cross-view training (CVT)~\cite{clark:2018:ICLR} considers the auxiliary loss by making a prediction from an unlabeled input with a restricted view.
On the other hand, in our \model{} framework, we do not use unlabeled data as additional training data for the baseline DNN.
Instead, we use the unlabeled data to train the \imitator{}s to imitate the baseline DNN.
The advantage of such usage is that one can choose an arbitrary architecture for the \imitator{}s.
In this study, we design the \imitator{} to be computationally simpler than the baseline DNN to ensure better scalability with the amount of unlabeled data (Table~\ref{table:speed}).

The idea of our \textit{expert-imitator} approach originated from the SSL framework proposed by~\citet{suzuki:2008:ACL}.
They incorporated several simple generative models as a set of additional features for a supervised linear conditional random field classifier.
Our \expert{} and \imitator{} can be regarded as their linear classifier and the generative models, respectively.
In addition, they empirically demonstrated that the performance has a linear relationship with the logarithm of the unlabeled data size.
We empirically demonstrate that the proposed method also exhibits similar behavior (Figure~\ref{fig:unlabeled}), namely, increasing the amount of unlabeled data reduces the error rate of the \expert{}.

One of the major SSL approaches in NLP is to pre-train a language model over unlabeled data.
The pre-trained weights have many uses, such as parameter initialization~\cite{dai:2015:NIPS} and  as a source of additional features~\cite{mccann:2017:NIPS,peters:2017:ACL,peters:2018:NAACL}, in downstream tasks.
For example,~\citet{peters:2018:NAACL} have recently trained a bi-directional LSTM language model using the One Billion Word Benchmark dataset~\cite{chelba:2014:interspeech}.
They utilized the hidden state of the LSTM as contextualized embedding, called \textit{ELMo} embedding, and achieved state-of-the-art results in many downstream tasks.
In our experiment, we empirically demonstrate that the proposed \mein{} is complementary to the pre-trained language model approach.
Specifically, we show that by combining the two approaches, we can further improve the performance of the baseline DNN.

\section{Task Description and Notation Rules}
This section gives a formal definition of the text classification task discussed in this paper.
Let $\mathcal{V}$ represent the vocabulary of the input sentences.
$\bm{x}_{t} \in \{0,1\}^{|\mathcal{V}|}$ denotes the one-hot vector of the $t$-th token (word) in the input sentence,
where $|\mathcal{V}|$ represents the number of tokens in $\mathcal{V}$.
Here, we introduce the short notation form $(\bm{x}_{t})_{t=1}^{T}$ to represent a sequence of vectors for simplicity, that is, $(\bm{x}_{t})_{t=1}^{T}=(\bm{x}_1,\dots,\bm{x}_{T})$.
Suppose we have an input sentence that consists of $T$ tokens.
For a succinct notation, we introduce $\bm{X}$ to represent a sequence of one-hot vectors that corresponds to the tokens in the input sentence, namely, $\bm{X}=(\bm{x}_{t})_{t=1}^{T}$.
$\mathcal{Y}$ denotes a set of output classes.
Let $y \in \{1,\dots,|\mathcal{Y}|\}$ be an integer that represents the output class ID. %
In addition, we define $\bm{X}_{a:b}$ as the subsequence of $\bm{X}$ from index $a$ to index $b$, namely, $\bm{X}_{a:b} = (\bm{x}_{a},\bm{x}_{a+1}\dots,\bm{x}_{b})$ and $1 \leq a \leq b \leq T$.
We also define $\bm{x}[i]$ as the $i$-th element of vector $\bm{x}$.
For example, if $\bm{x}=(5,2,1,-1)^{\top}$, then $\bm{x}[2]=2$ and $\bm{x}[4]=-1$.

In the supervised training framework for text classification tasks modeled by DNNs, we aim to maximize the (conditional) probability $p(y|\bm{X})$ over a given set of labeled training data $(\bm{X}, y) \in \mathcal{D}_{s}$ by using DNNs.
In the semi-supervised training, the objective of maximizing the probability is identical but we also use a set of unlabeled training data $\bm{X} \in \mathcal{D}_{u}$.

\section{Baseline Network: LSTM with MLP}
\label{sec:baseline}
In this section, we briefly describe a baseline DNN for text classification.
Among the many choices, we select the \textit{LSTM-based text classification model} described by~\citet{miyato:2017:ICLR} as our baseline DNN architecture since they achieved the current best results on several well-studied text classification benchmark datasets.
The network consists of the LSTM~\cite{hochreiter:1997:long} cell and a multi layer perceptron (MLP).

First, the LSTM cell calculates a hidden state sequence $(\bm{h}_{t})_{t=1}^{T}$, where $\bm{h}_{t} \in \mathbb{R}^{H}$ for all $t$ and $H$ is the size of the hidden state, as $\bm{h}_{t} = \LSTMU (\bm{E}\bm{x}_{t}, \bm{h}_{t-1})$.
Here, $\bm{E} \in \mathbb{R}^{D \times |\mathcal{V}|}$ is the word embedding matrix, $D$ denotes the size of the word embedding, and $\bm{h}_{0}$ is a zero vector.

Then the $T$-th hidden state $\bm{h}_{T}$ is passed through the MLP, which consists of a single fully connected layer with ReLU nonlinearity~\cite{glorot:2011:deep}, to compute the final hidden state $\bm{s} \in \mathbb{R}^{M}$.
Specifically, $\bm{s}$ is computed as $\bm{s} = \ReLU(\bm{W}_{h}\bm{h}_{T} + \bm{b}_{h})$,
where $\bm{W}_{h} \in \mathbb{R}^{M \times H}$ is a trainable parameter matrix and $\bm{b}_{h} \in \mathbb{R}^{M}$ is a bias term.
Here, $M$ denotes the size of the final hidden state of the MLP.

Finally, the baseline DNN estimates the conditional probability from the final hidden state $\bm{s}$ as follows:
\begin{align}
  z_{y} &= \tp{\bm{w}_{y}}\bm{s} + b_{y}, \label{eq:baselogit} \\
  p(y | \bm{X}, \bm{\Theta}) &= \frac{\exp(z_{y})}{\sum_{y^{\prime} \in \mathcal{Y}}\exp(z_{y^{\prime}})}\label{eq:softmax},
\end{align}
where $\bm{w}_{y} \in \mathbb{R}^{M}$ is the weight vector of class $y$ and $b_{y}$ is the scalar bias term of class $y$.
Also, $\bm{\Theta}$ denotes all the trainable parameters of the baseline DNN.

For the training process of the parameters in the baseline DNN $\bm{\Theta}$, we seek the (sub-)optimal parameters that minimize the (empirical) negative log-likelihood for the given labeled training data $\mathcal{D}_{s}$, which can be written as the following optimization problem:
\begin{align}
  \bm{\Theta}^{\prime} &= \argmin\limits_{\bm{\Theta}} \big\{ L_{s}(\bm{\Theta}| \mathcal{D}_{s}) \big\}, \label{eq:argmin1} \\
  L_{s}(\bm{\Theta}| \mathcal{D}_{s}) &= - \frac{1}{\vert \mathcal{D}_{s} \vert} \sum_{(\bm{X}, y) \in \mathcal{D}_{s}} \log \big( p(y | \bm{X}, \bm{\Theta}) \big),
 \label{eq:loss1}
\end{align}
where $\bm{\Theta}^{\prime}$ represents the set of obtained parameters in the baseline DNN, by solving the above minimization problem.
Practically, we apply a variant of a stochastic gradient descent algorithm such as Adam~\cite{kingma:2015:ICLR}.

\section{Proposed Model: Mixture of Expert/Imitator Networks (\model{})}

Figure \ref{fig:framework} gives an overview of the proposed method, which we refer to as \model{}.
\model{} consists of an expert network (\expert{}) and a set of imitator networks (\imitator{}s).
Once trained, the \expert{} and the set of \imitator{}s jointly predict the label of a given input $\bm{X}$.
Figure~\ref{fig:framework} shows the baseline DNN (LSTM with MLP) as an example of the \expert{}.
Note that \model{} can adopt an arbitrary classification network as the \expert{}.

\subsection{Basic Idea}
A brief description of \model{} is as follows:
(1) The \expert{} is trained using labeled training data.
Thus, the \expert{} is expected to be very accurate over inputs that are similar to the labeled training data.
(2) \imitator{}s (we basically assume that we have more than one \imitator{}) are trained to imitate the \expert{}.
To accomplish this, we train each \imitator{} to minimize the Kullback–Leibler (KL) divergence between estimations of label distributions of the \expert{} and the \imitator{}s over the unlabeled data.
(3) Our final classification network is a mixture of the \expert{} and \imitator{}(s).
Here, we fine-tune the \expert{} using the labeled training data jointly with the estimations of all the \imitator{}s.

The basic idea underlying \model{} is that we force each \imitator{} to imitate estimated label distributions with only a limited view of the given input.
Specifically, we adopt a sliding window to divide the input into several fragments of n-grams.
Given a large amount of unlabeled data and the estimation by the \expert{}, the \imitator{} learns to represent the label ``tendency'' of each fragment in the form of a label distribution (i.e., certain n-grams are more likely to have positive/negative labels than others).
Our assumption here is that this tendency can potentially contribute a set of features for the classification.
Thus, after training the \imitator{}s, we jointly optimize the \expert{} and the weight of each feature.
Here, \model{} may control the contribution of each feature by updating the corresponding weight.

Intuitively, our \model{} approach can be interpreted as a variant of several successful machine learning techniques for DNNs.
For example, \model{} shares the core concept with the mixture-of-experts technique (MoE)~\cite{jacobs:1991:neural,shazeer:2017:ICLR}.
The difference is that MoE considers a mixture of several \expert{}s, whereas \model{} generates a mixture from a single \expert{} and a set of \imitator{}s.
In addition, one can interpret \model{} as a variant of the ensemble, bagging, voting, or boosting technique since the \expert{} and the \imitator{}s jointly make a prediction.
Moreover, we train each \imitator{} by minimizing the KL-divergence between the \expert{} and the \imitator{} through unlabeled data.
This process can be seen as a form of ``knowledge distillation''~\cite{ba:2014:NIPS,hinton:2014:NIPS}.
We utilize these methodologies and formulate the framework as described below.
\begin{figure*}[t!]
    \center
    \includegraphics[width=0.9\hsize]{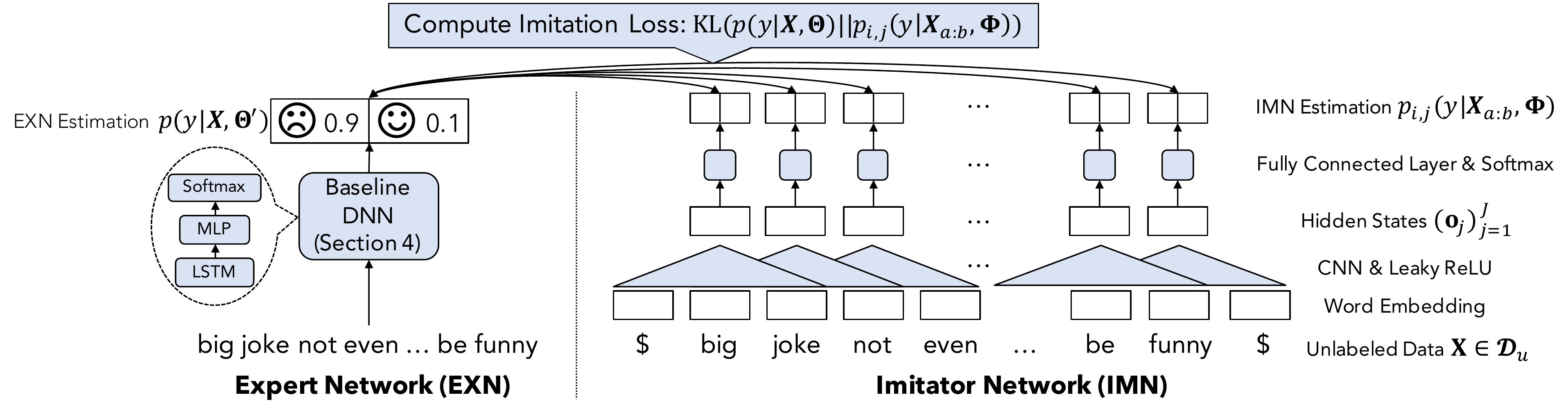}
    \caption{
    Overview of the 1st \imitator{} ($c_{1}=1$). The \imitator{} must predict the label estimation of the \expert{} from a limited amount of information.
    \$ denotes a special token used to pad the input (a zero vector).
    }
    \label{fig:ssm}
\end{figure*}

\subsection{Network Architecture}
\label{section:network}
Let $\sigma(\cdot)$ be the sigmoid function defined as $\sigma(\lambda) = (1+\exp(-\lambda))^{-1}$.
$\bm{\Phi}$ denotes a set of trainable parameters of the \imitator{}s and $I$ denotes the number of \imitator{}s.
Then, the \expert{} combined with a set of \imitator{}s models the following (conditional) probability:
\begin{align}
 &p(y|\bm{X}, \bm{\Theta}, \bm{\Phi},\bm{\Lambda}) = \frac{\exp(z_{y}^{\prime})}{\sum_{y^{\prime} \in \mathcal{Y}}\exp(z^{\prime}_{y^{\prime}})}\label{eq:prob},\\
&\mbox{where }\,\,\,
 z^{\prime}_y  = \displaystyle z_y
 + \sum_{i=1}^{I}\sigma(\lambda_{i})\bm{\alpha}_{i}[y]. \label{eq:logit}
\end{align}
$\lambda_{i}$ is a scalar parameter that controls the contribution of logit $\bm{\alpha}_{i}$ of the $i$-th \imitator{} and $\bm{\Lambda}$ is defined as $\bm{\Lambda} = \{\lambda_{1},\dots,\lambda_{I}\}$.
Here, logit $\bm{\alpha}_{i}$ represents an estimated label distribution, which we assume to be a feature.
Note that the first term of Equation~\ref{eq:logit} is the baseline DNN logit $z_{y}=\tp{\bm{w}_{y}}\bm{s} + {b}_y$ (Equation~\ref{eq:baselogit}).
In addition, if we set $\sigma(\lambda_{i})=0$ for all $i$, then Equation~\ref{eq:prob} becomes identical to Equation \ref{eq:softmax} regardless of the value of $\bm{\Phi}$.

$c_{i}$ denotes the window size of the $i$-th \imitator{}.
Given an input $\bm{X}$ and the $i$-th \imitator{}, we create $J$ inputs with a sliding window of size $c_{i}$.
Then the \imitator{} predicts the \expert{} for each input and generates $J$ predictions as a result.
We compute the $i$-th imitator logit $\bm{\alpha}_{i}$ by taking the average of these predictions.
Specifically, $\bm{\alpha}_{i}$ is defined as
\begin{align}
 \bm{\alpha}_{i} &= \log\biggr( \frac{1}{J} \sum_{j=1}^{J}p_{i,j}(y |\bm{X}_{a:b}, \bm{\Phi}) \biggr), \label{eq:submodel}\\
&\mbox{where} \quad a = j - c_{i} \quad \mbox{and} \quad b = j + c_{i}.  \nonumber
\end{align}
Here, $a$ is a scalar index that represents the beginning of the window.
Similarly, $b$ represents the last index of the window.

\subsection{Definition of \imitator{}s}
Note that the architecture of the \imitator{} used to model Equation~\ref{eq:submodel} is essentially arbitrary.
In this research, we adopt a single-layer CNN for modeling $p_{i,j}(y|\bm{X}_{a:b}, \bm{\Phi})$.
This is because a CNN has high computational efficiency~\cite{gehring:2017:ICML}, which is essential for our primary focus: scalability with the amount of unlabeled data.

Figure~\ref{fig:ssm} gives an overview of the architecture of the \imitator{}.
First, the \imitator{} takes a sequence of word embeddings of input $\bm{X}$ and computes a sequence of hidden states $(\bm{o}_j)_{j=1}^{J}$ by applying a \textit{one-dimensional convolution}~\cite{kalchbrenner:2014:ACL} and leaky ReLU nonlinearity~\cite{maas:2013:ICML}.
We ensure that $J$ is always equal to $T$.
To achieve this, we pad the beginning and the end of the input $\bm{X}$ with zero vectors $\bm{0} \in \mathbb{R}^{|\mathcal{V}^{\prime}| \times c_{i}}$, where $|\mathcal{V}^{\prime}|$ denotes the vocabulary size of the \imitator{}.

As explained in Section~\ref{section:network}, each \imitator{} has a predetermined and fixed window size $c_{i}$.
One can choose an arbitrary window size for the $i$-th \imitator{}.
Here, we define $c_{i}$ as $c_{i} = i$ for simplicity.
For example, as shown in Figure~\ref{fig:ssm}, the 1st \imitator{} ($i=1$) has a window size of $c_{1}=1$.
Such a network imitates the estimation of the \expert{} from three consecutive tokens.

Then the $i$-th \imitator{} estimates the probability $p_{i,j}(y |\bm{X}, \bm{\Phi})$ from each hidden state $\bm{o}_{j}$ as
\begin{align}
 p_{i, j}(y | \bm{X}_{a:b}, \bm{\Phi}) = \frac{\exp(\bm{w}^{\prime\top}_{i,y}\bm{o}_{j} + b_{i,y}^{\prime})}{\sum_{y^{\prime} \in \mathcal{Y}} \exp(\bm{w}^{\prime\top}_{i,y^{\prime}}\bm{o}_{j} + b_{i,y^{\prime}}^{\prime})},\label{eq:softmax2}
\end{align}
where $\bm{w}^{\prime}_{i,y} \in \mathbb{R}^{N}$ is the weight vector of the $i$-th \imitator{} and $b_{i,y}^{\prime}$ is the scalar bias term of class $y$.
$N$ denotes the CNN kernel size.

\begin{algorithm}[t!]
  \DontPrintSemicolon
  \KwData{Labeled data $\mathcal{D}_{s}$ and unlabeled data $\mathcal{D}_{u}$}

  \KwResult{Trained set of parameters $\widehat{\bm{\Theta}}, \widehat{\bm{\Phi}}, \widehat{\bm{\Lambda}}$}
  $\bm{\Theta}^{\prime}\gets\argmin\limits_{\bm{\Theta}}\{L_{s}(\bm{\Theta}|\mathcal{D}_s)\}$
  \hspace{\fill}{$\triangleright$ {\scriptsize Train \expert{} (Equation~\ref{eq:argmin1})}}

  $\widehat{\bm{\Phi}} \gets\argmin\limits_{\bm{\Phi}}\{L_{u}(\bm{\Phi} | \bm{\Theta}^{\prime}, \mathcal{D}_u)\}$ \hspace{\fill}{$\triangleright$ {\scriptsize Train \imitator{}(s) (Equation~\ref{eq:nolabelloss})}}

  $\widehat{\bm{\Theta}}, \widehat{\bm{\Lambda}}\gets\argmin\limits_{\bm{\Theta},\bm{\Lambda}}\{\!L_{s}^{\prime}(\bm{\Theta}, \bm{\Lambda} | \widehat{\bm{\Phi}},\!\mathcal{D}_s\!)\!\}$
  \hspace{\fill}{$\triangleright$ {\scriptsize Train \expert{} (Equation~\ref{eq:loss_final})}}
	\caption{Training framework of \model{}}
  \label{alg:training}
\end{algorithm}

\subsection{Training Framework}
First, we define the \textit{imitation loss} of each \imitator{} as the KL-divergence between the estimations of the label distributions of the \expert{} and the \imitator{} given (unlabeled) data $\bm{X}$, namely, $\textrm{KL} (p(y|\bm{X}, \bm{\Theta} ) || p_{i,j}(y | \bm{X}_{a:b}, \bm{\Phi}) )$.
Note that this imitation loss is defined for an input with the sliding window $\bm{X}_{a:b}$.
Thus, this definition effectively accomplishes the concept, i.e., the \imitator{} making a prediction $p_{i,j}(y | \bm{X}_{a:b}, \bm{\Phi})$ from only a limited view of the given input $\bm{X}_{a:b}$.

Next, our objective is to estimate the set of optimal parameters by minimizing the negative log-likelihood of Equation~\ref{eq:prob} while also minimizing the total imitation losses for all \imitator{}s as biases of the network.
Therefore, we jointly solve the following two minimization problems for the parameter estimation of \model{}:
\begin{align}
 \widehat{\bm{\Theta}}, \widehat{\bm{\Lambda}}
 &= \argmin\limits_{\bm{\Theta},\bm{\Lambda}}\{L^{\prime}_{s}(\bm{\Theta}, \bm{\Lambda} | \widehat{\bm{\Phi}}, \mathcal{D}_s)\}
 \label{eq:joint1}
 \\
 \widehat{\bm{\Phi}}
 &= \argmin\limits_{\bm{\Phi}}\{L_{u}(\bm{\Phi} | \bm{\Theta}', \mathcal{D}_u)\}
 \label{eq:joint2}
 .
\end{align}
As described in Equations~\ref{eq:joint1} and~\ref{eq:joint2},
we update the different sets of parameters depending on the labeled/unlabeled training data.
Specifically, we use the labeled training data $(\bm{X}, y) \in \mathcal{D}_s$ to update the set of parameters in the \expert{}, $\bm{\Theta}$, and the set of mixture parameters of the \imitator{}s, $\bm{\Lambda}$.
In addition,  we use the unlabeled training data $\bm{X} \in \mathcal{D}_{u}$ to update the parameters of the \imitator{}s, $\bm{\Phi}$.

To ensure an efficient training procedure, the training framework of \model{} consists of three consecutive steps (Algorithm~\ref{alg:training}).
First, we perform standard supervised learning to obtain $\bm{\Theta}'$ using labeled training data while keeping ${\lambda}_i=-\infty$ unchanged for all $i$ during the training process to ensure that $\sigma(\lambda_i)=0$ in Equation~\ref{eq:logit}.
Note that this optimization step is essentially equivalent to that of the baseline DNN (Equation~\ref{eq:loss1}).

Second, we estimate the set of \imitator{} parameters $\bm{\Phi}$ by solving the minimization problem in Equation~\ref{eq:joint2} with the following loss function: %
\begin{align}
  L_{u}(\bm{\Phi}| \bm{\Theta}', \mathcal{D}_u) =& \frac{1}{\vert \mathcal{D}_{u} \vert} \! \sum_{\bm{X} \in \mathcal{D}_{u}} \! \sum_{i=1}^{I}  \sum_{j=1}^{J} \textrm{KL} (p || p_{i,j} ), \label{eq:nolabelloss}\\
  \textrm{KL} (p || p_{i,j} )=& - \sum_{y\in{\cal Y}}  p(y|\bm{X}, \bm{\Theta}' ) \log \big(p_{i,j}(y | \bm{X}_{a:b}, \bm{\Phi}) \big) \nonumber \\ \quad &+ const, \label{eq:loss1A}
\end{align}
where $\textrm{KL} (p || p_{i,j} )$ is a shorthand notation of the imitation loss $\textrm{KL} (p(y|\bm{X}, \bm{\Theta} ) || p_{i,j}(y | \bm{X}_{a:b}, \bm{\Phi}) )$ and $const$ is a constant term that is independent of $\bm{\Phi}$.

Finally, we estimate $\bm{\Theta}$ and $\bm{\Lambda}$ by solving the minimization problem in Equation~\ref{eq:joint1} with the following loss function:
\begin{align}
 L^{\prime}_{s}(\bm{\Theta},\bm{\Lambda}| \widehat{\bm{\Phi}}, \mathcal{D}_s ) &= - \frac{1}{\vert \mathcal{D}_{s} \vert} \sum_{(\bm{X}, y) \in \mathcal{D}_{s}}
 \!\!\log \Big(\!p(y | \bm{X}, \bm{\Theta}, \widehat{\bm{\Phi}},\!\bm{\Lambda})\!\Big). \label{eq:loss_final}
\end{align}

\section{Experiments}
\label{section:experiment}
To investigate the effectiveness of \model{}, we conducted experiments on two text classification tasks: (1) a sentiment classification (SEC) task and (2) a category classification (CAC) task.

\subsection{Datasets}
For SEC, we selected the following widely used benchmark datasets: IMDB~\cite{maas:2011:ACL}, Elec~\cite{johnson:2015:NIPS}, and Rotten Tomatoes (Rotten)~\cite{pang:2005:ACL}.
For the Rotten dataset, we used the Amazon Reviews dataset~\cite{mcauley:2013:ACM} as unlabeled data, following previous studies~\cite{dai:2015:NIPS,miyato:2017:ICLR,sato:2018:IJCAI}.
For CAC, we used the RCV1 dataset~\cite{lewis:2004:RNB}.
Table~\ref{table:dataset} summarizes the characteristics of each dataset\footnote{DBpedia~\cite{lehmann:2015:SW} is another widely adopted CAC dataset. We did not use this dataset in our experiment because it does not contain unlabeled data.}.

\begin{table}[t!]
 \centering
 \small
 \tabcolsep 4pt
   \begin{tabular}{llrrrrr}
    \toprule
       Task                 & Dataset      & Classes &  Train & Dev & Test & Unlabeled \\ \midrule
      \multirow{3}{*}{SEC}  & Elec         & 2 &  22,500  &  2,500 &   25,000  & 200,000  \\
                            & IMDB         & 2 &  21,246  &  3,754 &   25,000  &  50,000  \\
                            & Rotten       & 2 &  8,636   &  960  &  1,066   & 7,911,684   \\\midrule
       CAC                  & RCV1         & 55 &  14,007 &  1,557 &    49,838 &  668,640      \\
   \bottomrule
   \end{tabular}
 \caption{Summary of datasets. Each value represents the number of instances contained in each dataset.}
   \label{table:dataset}
\end{table}

\subsection{Baseline DNNs}
In order to investigate the effectiveness of the \model{} framework, we combined the \imitator{} with following three distinct \expert{}s and evaluated their performance:
\begin{itemize}
  \item \weakbase{}: This is the baseline DNN (LSTM with MLP) described in Section~\ref{sec:baseline}.
  \item \strongbase{}: Following \citet{dai:2015:NIPS}, we initialized the embedding layer and the LSTM with a pre-trained RNN-based language model (LM)~\cite{bengio:2003:neural}.
  We trained the language model using the labeled training data and unlabeled data of each dataset.
  Several previous studies have adopted this network as a baseline~\cite{miyato:2017:ICLR,sato:2018:IJCAI}.
  \item \adv{}: Adversarial training (ADV)~\cite{goodfellow:2015:ICLR} adds small perturbations to the input  and makes the network robust against noise. \citeauthor{miyato:2017:ICLR}~\shortcite{miyato:2017:ICLR} applied ADV to \strongbase{} for a text classification. We used the reimplementation of their network.
\end{itemize}
Note that these three \expert{}s have an identical network architecture, as described in Section~\ref{sec:baseline}.
The only difference is in the initialization or optimization strategy of the network parameters.

To the best of our knowledge, \adv{} provides a performance competitive with the current best result for the configuration of supervised learning (using labeled training data only).
Thus, if the \imitator{} can improve the performance of a strong baseline, the results will strongly indicate the effectiveness of our method.
\subsection{Network Configurations}
Table~\ref{table:hyperparams} summarizes the hyperparameters and network configurations of our experiments.
We carefully selected the settings commonly used in the previous studies~\cite{dai:2015:NIPS,miyato:2017:ICLR,sato:2018:IJCAI}.

We used a different set of vocabulary for the \expert{} and the \imitator{}s.
We created the \expert{} vocabulary $\mathcal{V}$ by following the previous studies~\cite{dai:2015:NIPS,miyato:2017:ICLR,sato:2018:IJCAI}, i.e., we removed the tokens that appear only once in the whole dataset.
We created the \imitator{} vocabulary $\mathcal{V}^{\prime}$ by byte pair encoding (BPE)~\cite{sennrich:2016:ACL}\footnote{We used sentencepiece~\cite{kudo:2018:EMNLP} (\url{https://github.com/google/sentencepiece}) for the BPE operations.}.
The BPE merge operations are jointly learned from the labeled training data and unlabeled data of each dataset.
We set the number of BPE merge operations to 20,000.

\begin{table}[t!]
 \centering
 \small
 \tabcolsep 1pt
    \begin{tabular}{clc}
     \toprule
       & Hyperparameter          &  Value \\ \midrule

       \multirow{6}{*}{
       \begin{tabular}{c}
        \expert{} \\
        (baseline DNN)
        \end{tabular}
       }
              & Word Embedding Dim. ($D$)       & 256 \\
              & Embedding Dropout Rate          & 0.5  \\
              & LSTM Hidden State Dim. ($H$)    & 1024  \\
              & MLP Dim. ($M$) for SEC Task     & 30 \\
              & MLP Dim. ($M$) for CAC Task     & 128  \\
              & Activation Function             & ReLU  \\ \midrule
      \multirow{4}{*}{\imitator{}}
              & CNN Kernel Dim. ($N$)           & 512 \\
              & Word Embedding Dim.             & 512   \\
              & Activation Function             & Leaky ReLU   \\
              & Number of \imitator{}s ($I$)    & 4   \\ \midrule
      \multirow{7}{*}{Optimization}
              & Algorithm                       & Adam \\
              & Mini-Batch Size                 & 32 \\
              & Initial Learning Rate           & 0.001 \\
              & Fine-tune Learning Rate         & 0.0001 \\
              & Decay Rate                      & 0.9998 \\
              & Baseline Max Epoch              & 30 \\
              & Fine-tune Max Epoch             & 30 \\
    \bottomrule
   \end{tabular}
 \caption{Summary of hyperparameters}
   \label{table:hyperparams}
\end{table}

\subsection{Results}
\label{section:result}
\begin{table}[t!]
 \centering
 \small
 \tabcolsep 1pt
\begin{tabular}{lrrrrr}
\toprule
Method                                              & \multicolumn{1}{c}{Elec} & \multicolumn{1}{c}{IMDB} & \multicolumn{1}{c}{Rotten} & \multicolumn{1}{c}{RCV1} \\ \midrule
\weakbase{}                                         & 10.09 & 10.98  & 26.47  & 14.14 \\
\weakbase{}+\imitator{} (Random)\unlabeled{}        & 9.87  & 10.75  & 27.27  & 14.04 \\
\textbf{\weakbase{}+\imitator{}}\unlabeled{}                 & \textbf{8.83}  & \textbf{10.04}  & \textbf{24.93}  & \textbf{12.31} \\\midrule
\strongbase{}\unlabeled{}                           & 5.72  & 7.25   & 16.80  & 8.37  \\
\strongbase{}+\imitator{} (Random)\unlabeled{}      & 5.71  & 7.01   & 16.78  & 7.83  \\
\textbf{\strongbase{}+\imitator{}}\unlabeled{}               & \textbf{5.48}  & \textbf{6.51}   & \textbf{15.91}  & \textbf{7.53}  \\\midrule
\adv{}\unlabeled{}                                  & 5.38  & 6.58   & 15.73  & 7.89  \\
\adv{}+\imitator{} (Random)\unlabeled{}             & 5.34  & 6.27   & 15.11  & 7.78  \\
\textbf{\adv{}+\imitator{}}\unlabeled{}                      & \textbf{5.14}\significant  & \textbf{6.07}\significant   & \textbf{13.98}  & \textbf{7.51}\significant  \\
\midrule\midrule
\vat{} (rerun) \unlabeled{}                         & 5.47  & 6.20   & 18.50  & 8.44  \\
\vat{} (\texttt{Miyato 2017})\unlabeled{}           & 5.54  & 5.91   & 19.1~~~& 7.05  \\
\vat{} (\texttt{Sato 2018})\unlabeled{}             & 5.66  & 5.69   & 14.26  & 11.80  \\
\ivat{} (\texttt{Sato 2018})\unlabeled{}            & 5.18  & 5.66   & 14.12  & 11.68 \\
\bottomrule
\end{tabular}
 \caption{Test performance (error rate (\%)) on each dataset. \textbf{A lower error rate indicates better performance.} Models using the unlabeled data are marked with $\dag$. Results marked with ${}^{*}$ are statistically significant compared with \adv{}. \texttt{Miyato 2017}: the result reported by~\citet{miyato:2017:ICLR}. \texttt{Sato 2018}: the result reported by~\citet{sato:2018:IJCAI}.
 }
 \label{table:result}
\end{table}

Table~\ref{table:result} summarizes the results on all benchmark datasets, where the evaluation metric is the error rate.
Therefore, a lower value indicates better performance.
Here, all the reported results are \textbf{the average of five distinct trials} using five different random seeds.
Moreover, for each trial, we automatically selected the best network in terms of the performance on the validation set among the networks obtained at every epoch.
For comparison, we also performed experiments on training baseline DNNs (\weakbase{}, \strongbase{}, and \adv{}) with incorporating random vectors as the replacement of \imitator{}s, which is denoted as ``+\imitator{}~(Random)''.
Moreover, we present the published results of \vat{}~\cite{miyato:2017:ICLR} and \ivat{}~\cite{sato:2018:IJCAI} in the bottom three rows of Table~\ref{table:result}, which are the current state-of-the-art networks that adopt unlabeled data.
For \vat{}, we also report the result of the reimplemented network, denoted as ``\vat{} (rerun)''.

As shown in Table~\ref{table:result}, incorporating the \imitator{}s consistently improved the performance of all baseline DNNs across all benchmark datasets.
Note that the source of these improvements is not the extra set of parameters $\bm{\Lambda}$ but the outputs of the \imitator{}s.
We can confirm this fact by comparing the results of \imitator{}s, ``+\imitator{}'', with those of random vectors, ``+\imitator{} (Random)'', since the difference between these two settings is the incorporation of \imitator{}s or random vectors.

The most noteworthy observation about \model{} is that the amount of the improvement upon incorporating the \imitator{} is nearly consistent, regardless of the performance of the base \expert{}.
For example, Table~\ref{table:result} shows that the \imitator{} reduced the error rates of \weakbase{}, \strongbase{}, and \adv{} by 1.54\%, 0.89\%, and 1.22\%, respectively, for the Rotten dataset.
From these observations, the \imitator{} has the potential to further improve the performance of much stronger \expert{}s developed in the future.

We also remark that our best configuration, \adv{}+\imitator{}, outperformed \vat{} (rerun) on all datasets%
\footnote{The performance of our \vat{} (rerun) is lower than the performances reported by \citet{miyato:2017:ICLR} except for the Elec and Rotten datasets.
Through extensive trials to reproduce their results, we found that the hyperparameter of the RNN language model is extremely important in determining the final performance; therefore, the strict reproduction of the published results is significantly difficult.
In fact, a similar difficulty can be observed in Table~\ref{table:result}, where \vat{} (\texttt{Sato 2018}) has lower performance than \vat{} (\texttt{Miyato 2017}) on the Elec and RCV1 datasets.
Thus, we believe that \vat{} (rerun) is the most reliable result for the comparison.
}.
In addition, the best configuration outperformed the current best published results on the Elec and Rotten datasets, establishing new state-of-the-art results.

As a comparison with the current strongest SSL method, we combined the \imitator{} with the current state-of-the-art VAT method, namely, \vat{}+\imitator{}.
In the Elec dataset, the \imitator{} improved the error rate from 5.47\% to 5.16\%.
This result indicates that the \imitator{} and VAT have a complementary relationship.
Note that utilizing VAT is challenging in terms of the scalability with the amount of unlabeled data.
However, if sufficient computing resources exist, then VAT and the \imitator{} can be used together to achieve even higher performance.

\section{Analysis}

\subsection{More Data, Better Performance Property}
We investigated whether the \model{} framework has the \textit{more data, better performance} property for unlabeled data.
Ideally, \model{} should achieve better performance by increasing the amount of unlabeled data.
Thus, we evaluated the performance while changing the amount of unlabeled data used to train the \imitator{}.

We selected the Elec and RCV1 datasets as the focus of this analysis.
We created the following subsamples of the unlabeled data for each dataset: \{5K, 20K, 50K, 100K, Full Data\} for Elec and \{5K, 50K, 250K, 500K, Full Data\} for RCV1.
In addition, for the Elec dataset, we sampled extra unlabeled data from the electronics section of the Amazon Reviews dataset~\cite{mcauley:2013:ACM} and constructed \{2M, 4M, 6M\} unlabeled data\footnote{We discarded instances from the unlabeled data when the non stop-words overlap with instances in the Elec test set. Thus, the unlabeled data and the Elec test set had no instances in common.}.
For each (sub)sample, we trained \adv{}+\imitator{} as explained in Section~\ref{section:experiment}.

\begin{figure}[t!]
\centering
\subcaptionbox{Elec\label{fig:elec}}{\includegraphics[width=0.49\columnwidth]{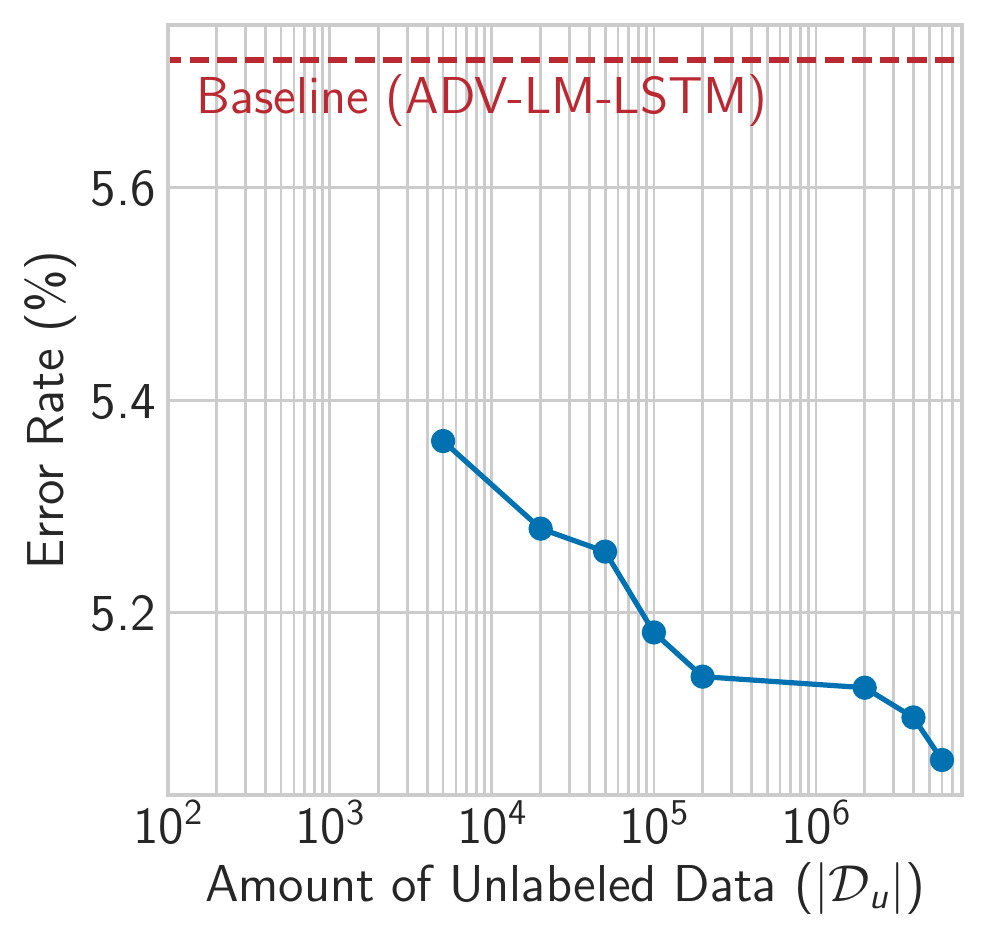}}
\subcaptionbox{RCV1\label{fig:rcv1}}{\includegraphics[width=0.49\columnwidth]{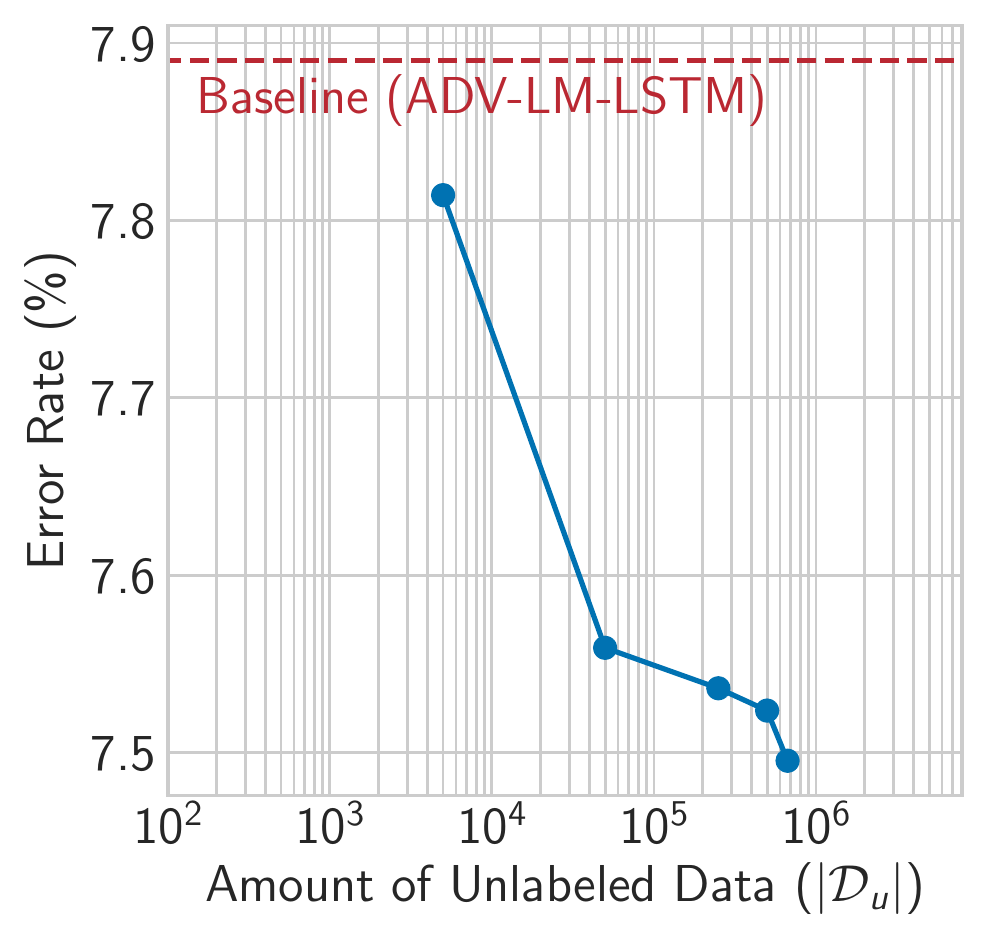}}
\caption{Error rate (\%) at different amounts of unlabeled data. The x-axis is in log-scale. \textbf{A lower error rate indicates better performance.} The dashed horizontal line represents the performance of the base \expert{} (\adv{}).}
\label{fig:unlabeled}
\end{figure}

Figures~\ref{fig:elec} and~\ref{fig:rcv1} demonstrate that increasing the amount of unlabeled data improved the performance of the \expert{}.
It is noteworthy that in Figure~\ref{fig:elec}, \adv{}+\imitator{} trained with 6M data achieved an error rate of 5.06\%, outperforming the best result in Table~\ref{table:result} (5.14\%).
These results explicitly demonstrate the \textit{more data, better performance} property of the \model{} framework.
We also report that the training process on the largest amount of unlabeled data (6M) only took approximately a day.

\subsection{Scalability with Amount of Unlabeled Data}
The primary focus of the \model{} framework is its scalability with the amount of unlabeled data.
Thus, in this section, we compare the computational speed of the \imitator{}s with that of the base \expert{}.
We also compare the \imitator{}s with the state-of-the-art SSL method, \vat{}, and discuss their scalability.
Here, we focus on the computation in the training phase of the network, where the network processes both forward and backward computations.

We measured the number of tokens that each network processes per second.
We used identical hardware for each measurement, namely, a single NVIDIA Tesla V100 GPU.
We used the cuDNN implementation for the LSTM cell since it is highly optimized and substantially faster than the naive implementation~\cite{bradbury:2017:ICLR}.

\begin{table}[t!]
  \centering
  \small
\begin{tabular}{lrc}
\toprule
Method      & Tokens/sec & Relative Speed \\
\midrule
\strongbase{}                  & 41,914    &  -      \\\midrule
\adv{}                         & 13,791    &  0.33x  \\
\vat{}                         & 9,602     &  0.23x  \\
\imitator{} ($c_i=1$)          & 555,613   &  13.26x~~  \\
\imitator{} ($c_i=1,2$)        & 236,065   &  5.63x  \\
\imitator{} ($c_i=1,2,3$)      & 122,076   &  2.91x  \\
\imitator{} ($c_i=1,2,3,4$)    & 75,393    &  1.80x  \\
\bottomrule
\end{tabular}
\caption{Number of tokens processed per second during the training}
\label{table:speed}
\end{table}

Table~\ref{table:speed} summarizes the results.
The table shows that even the slowest \imitator{} ($c_i=1,2,3,4$) was 1.8 times faster than the optimized cuDNN LSTM network and eight times faster than \vat{}.
This indicates that it is possible to use an even larger amount of unlabeled data in a practical time to further improve the performance of the \expert{}.
In addition, note that each \imitator{} can be trained in \textit{parallel}.
Thus, if multiple GPUs are available, the training can be carried out much faster than reported in Table~\ref{table:speed}.

\subsection{Effect of Window Size of the \imitator{}}
\label{section:windowsize}
\begin{figure}[t!]
    \center
    \includegraphics[width=\hsize]{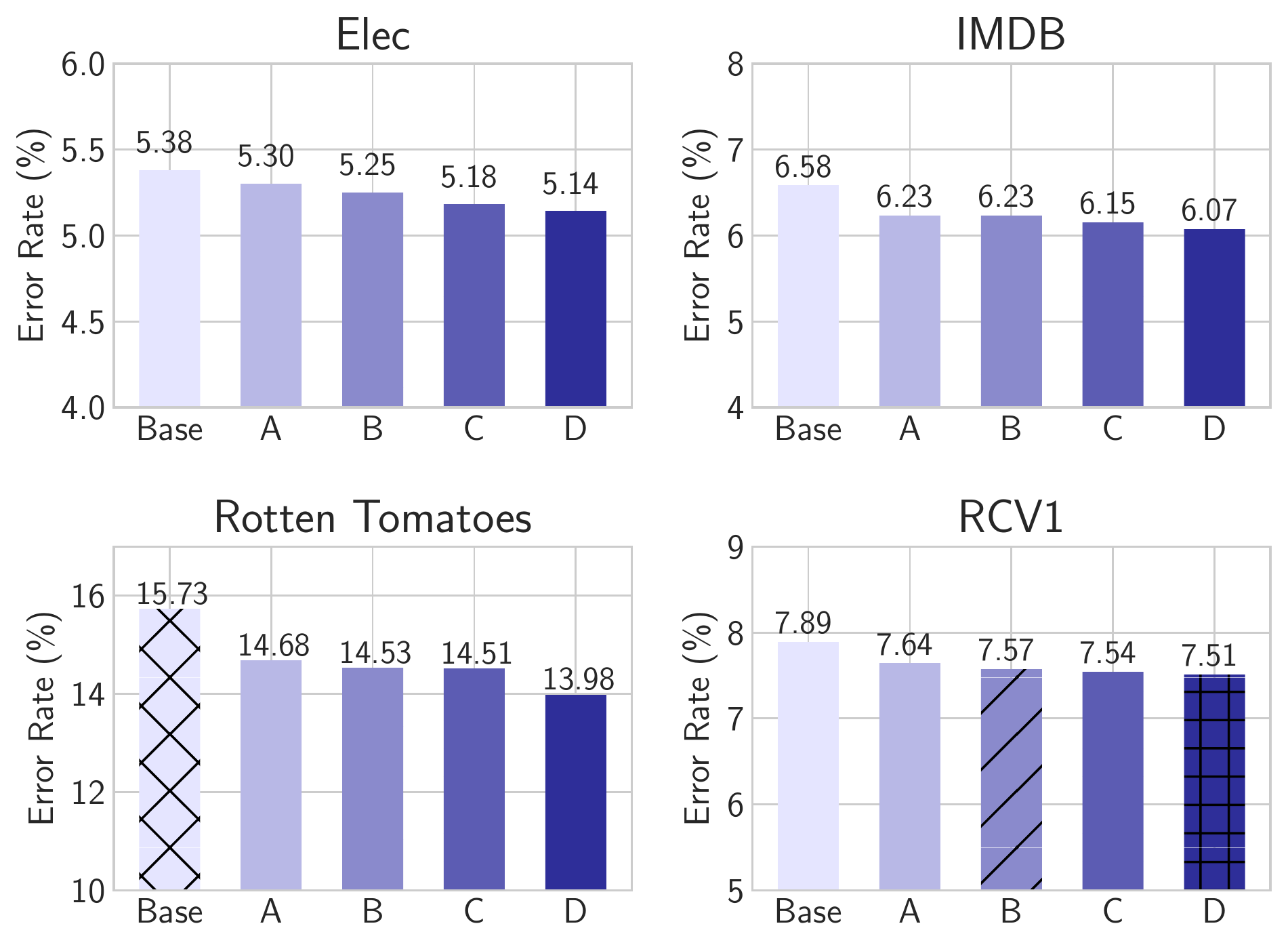}
    \caption{Effect of the \imitator{} with different window sizes $c_{i}$ on the final error rate (\%) of \adv{}. \textbf{A lower error rate indicates better performance.} \textbf{Base}: \expert{} (\adv{}) without the \imitator{}, \textbf{A}: $c_{i}=1$, \textbf{B}: $c_{i}=1,2$, \textbf{C}: $c_{i}=1,2,3$, \textbf{D}: $c_{i}=1,2,3,4$.}
    \label{fig:numberadv}
\end{figure}
In this section, we investigate the effectiveness of combining \imitator{}s with different window sizes $c_{i}$ on the final performance of the \expert{}.
Figure~\ref{fig:numberadv} summarizes the results across all datasets.
The figure shows that integrating an \imitator{} with a greater window size consistently reduced the error rate, and the \imitator{} with the greatest window size (\textbf{D}: $c_{i}=1,2,3,4$) achieved the best performance.
This observation implies that the context, which is captured by a greater window size, contributes to the performance.

\section{Discussion}
\subsection{Variations of the \imitator{}}

In this section, we discuss two possible variations of the \imitator{} to better understand its effectiveness in the \mein{} framework.

\subsubsection{Incorporating \imitator{} with Greater Window Size}
As discussed in Section~\ref{section:windowsize}, Figure~\ref{fig:numberadv} demonstrates that increasing the window size of the \imitator{} consistently improves the performance.
From this observation, one may hypothesize that integrating an \imitator{} with an even greater window size will be beneficial.
Thus, we carried out an experiment with such a configuration, i.e., $c_{i}=1,2,3,4,5$, and found that the hypothesis is valid.
For example, the error rates of \adv{}+\imitator{} ($c_{i}=1,2,3,4,5$) were 5.12\% and 6.00\% for Elec and IMDB, respectively, which are better than the values reported in Table~\ref{table:result}.

However, we found that a large window size has a major drawback; the training of \imitator{}s becomes significantly slower.
This undesirable property must be avoided as our primary focus is the scalability with the amount of unlabeled data.
Thus, we do not report these values as the main results of the experiment in Table~\ref{table:result}.

\subsubsection{Removing \imitator{}s with Smaller Window Sizes}

\begin{table}[]
\centering
\begin{tabular}{lr}
\toprule
Window Size     & Error Rate (\%) \\
\midrule
$c_{i}=1,2,3,4$ & 5.14            \\
$c_{i}=2,3,4$   & 5.18            \\
$c_{i}=3,4$     & 5.26            \\
$c_{i}=4$       & 5.23            \\
\bottomrule
\end{tabular}
\caption{Effect of removing \imitator{}s with smaller window sizes on the error rate (\%) of \adv{} on the Elec dataset. \textbf{A lower error rate indicates better performance.}}
\label{table:imncomb}
\end{table}

We also investigated the effectiveness of utilizing \imitator{}s with smaller window size in addition to the larger window sizes.
Table~\ref{table:imncomb} gives the results of this investigation, and we can see that combining \imitator{}s with smaller window sizes works better than incorporating a single \imitator{} with the greatest window size.

\subsection{Stronger Baseline DNN}
In this section, we discuss the results of two attempts to improve the performance of baseline DNNs.

\subsubsection{Increasing Number of Parameters}
The most straightforward means of improving the performance of baseline DNNs is to increase the number of parameters.
Thus, we doubled the word embedding dimension and trained \adv{}, namely, the \adv{}-Large model.
This model has approximately the same number of parameters as the \adv{}+\imitator{}.
However, the performance did not improve from that of the original \adv{}.
Specifically, the error rate degraded by 0.08 points for the IMDB dataset and was unchanged for the Elec dataset.

\subsubsection{Combining ELMo}
ELMo~\cite{peters:2018:NAACL} is one of the strongest SSL approaches in the research field.
Thus, we conducted an experiment with a baseline that utilizes ELMo.
Specifically, we combined \weakbase{} with the ELMo embeddings, namely, \elmo{}\footnote{We used the implementation available in AllenNLP~\cite{gardner:2017:AllenNLP}.}.
The error rate of this network on the IMDB test set was $8.67\%$, which is worse than that of \strongbase{} reported in Table~\ref{table:result}.
This result suggests that, at least in this task setting, pre-training the RNN language model for initialization is more effective than using the ELMo embeddings.

\section{Conclusion}
In this paper, we proposed a novel method for SSL, which we named Mixture of Expert/Imitator Networks (\model{}).
The \model{} framework consists of a baseline DNN, i.e., an \expert{}, and several auxiliary networks, \imitator{}s.
The unique property of our method is that the \imitator{}s learn to ``imitate'' the estimated label distribution of the \expert{} over the unlabeled data with only a limited view of the given input.
In this way, the \imitator{}s effectively learn a set of features that potentially contributes to improving the classification performance of the \expert{}.

Experiments on text classification datasets demonstrated that the \model{} framework consistently improved the performance of three distinct settings of the \expert{}.
We also trained the \imitator{}s with extra large-scale unlabeled data and achieved a new state-of-the-art result.
This result indicates that our method has the \textit{more data, better performance} property.
Furthermore, our method operates eight times faster than the current strongest SSL method (VAT), and thus, it has promising scalability to the amount of unlabeled data.

\bibliography{references}
\bibliographystyle{aaai}

\clearpage
\onecolumn
\appendix
\section*{Appendix}
\subsection*{Notation Rules and Tables}
This paper uses the following notation rules:
\begin{enumerate}
 \item Calligraphy letter represents a mathematical set (e.g., $\mathcal{D}_{s}$ denotes a set of labeled training data)
 \item Bold capital letter represents a (two-dimensional) matrix (e.g., $\bm{W}$ denotes a trainable matrix)
 \item Bold lower case letter represents a (one-dimensional) vector (e.g., $\bm{x}$ is a one-hot vector)
 \item Non-bold capital letter represents a fixed scalar value (e.g., $H$ denotes the LSTM hidden state dimension)
 \item Non-bold lower case letter represents a scalar variable (e.g., $t$ denotes a scalar time step $t$)
 \item Greek bold capital letter represents a set of (trainable) parameters (e.g., $\bm{\Theta}$ denotes a set of parameter of the \expert{})
 \item Non-bold Greek letter represents a scalar (trainable) parameter (e.g., $\lambda_{i}$ denotes a scalar trainable parameter)
 \item $(\bm{x}_{t})_{t=1}^{T}$ is a short notation for $(\bm{x}_1,\dots,\bm{x}_{T})$
 \item $\bm{x}[i]$ represents $i$-th element of the vector $\bm{x}$
 \item $\bm{X}_{a:b}$ represents an operation that slices a sequence of vectors $(\bm{x}_{a},\bm{x}_{a+1}\dots,\bm{x}_{b-1},\bm{x}_{b})$ from the matrix $\bm{X}$.
\end{enumerate}
A set of notations used in this paper is summarized in Table~\ref{table:notation}.
\begin{table*}[h!]
  \centering
  \small
  \begin{tabular}{cl|cl}
    \toprule
    Symbol                 & Description                                           & Symbol            & Description \\ \midrule
    $\bm{X}$               & sequence of one-hot vectors $\bm{x}$                  & $p$               & probability \\
    $\bm{x}$               & a one-hot vector representing single token (word)     & $\bm{W}_{h}$      & weight matrix for the LSTM hidden state $\bm{h}_{T}$ \\
    $\mathcal{Y}$          & set of output classes                                 & $\bm{\alpha}_{i}$ & $i$-th \imitator{} logit \\
    $y$                    & scalar class ID of the output                         & $L$               & loss function \\
    $\lambda_{i}$          & coefficient of $i$-th \imitator{}'s logit             & $\mathrm{KL}$     & KL-divergence function \\
    $\sigma$               & sigmoid function                                      & $\bm{w}_{y}$      & weight vector of softmax classifier for class $y$  \\
    $t$                    & variable for time step                                & $\bm{\Phi}$       & set of parameter of the \imitator{}\\
    $T$                    & time (typically denotes the sequence length)          & $\bm{\Theta}$     & set of parameter of the \expert{}  \\
    $\mathcal{V}$          & vocabulary of the baseline DNN                        & $\bm{\Lambda}$    & set of parameter for combining the \expert{} and \imitator{}(s) \\
    $\mathcal{V}^{\prime}$ & vocabulary of the \imitator{}                         & $J$               & number of outputs from a single \imitator{} \\
    $\mathcal{D}_s$        & set of labeled training data                          & $c_i$             & window size of the $i$-th \imitator{} \\
    $\mathcal{D}_u$        & set of unlabeled training data                        & $\bm{0}$          & concatenation of zero vector  \\
    $I$                    & number of \imitator{}s                                & $\bm{s}$          & MLP final hidden state\\
    $i$                    & general variable                                      & $\bm{o}_{j}$      & hidden state of the \imitator{}  \\
    $j$                    & general variable                                      & $a$               & start-index of sliding window  \\
    $\bm{E}$               & word embedding matrix                                 & $b$               & end-index of sliding window  \\
    $\bm{h}_{i}$           & LSTM hidden state                                     & $H$               & LSTM hidden state dimension   \\
    $D$                    & word embedding dimension of expert                    & $N$               & CNN Kernel dimension  \\
    $M$                    & MLP final hidden state dimension                      & $z_y$             & the \expert{} logit       \\
    $\bm{b}_{h}$           & a vector bias term for LSTM hidden state $\bm{h}_{T}$ & $z^{\prime}_{y}$  & \expert{} + \imitator{} logit   \\
    $b_y$                  & a scalar bias term for class $y$                      & -                 & - \\
    \bottomrule
  \end{tabular}
  \caption{Notation Table}
  \label{table:notation}
\end{table*}

\subsection*{Effect of Window Size of the \imitator{}}
\begin{figure}[t!]
    \center
    \includegraphics[width=\hsize]{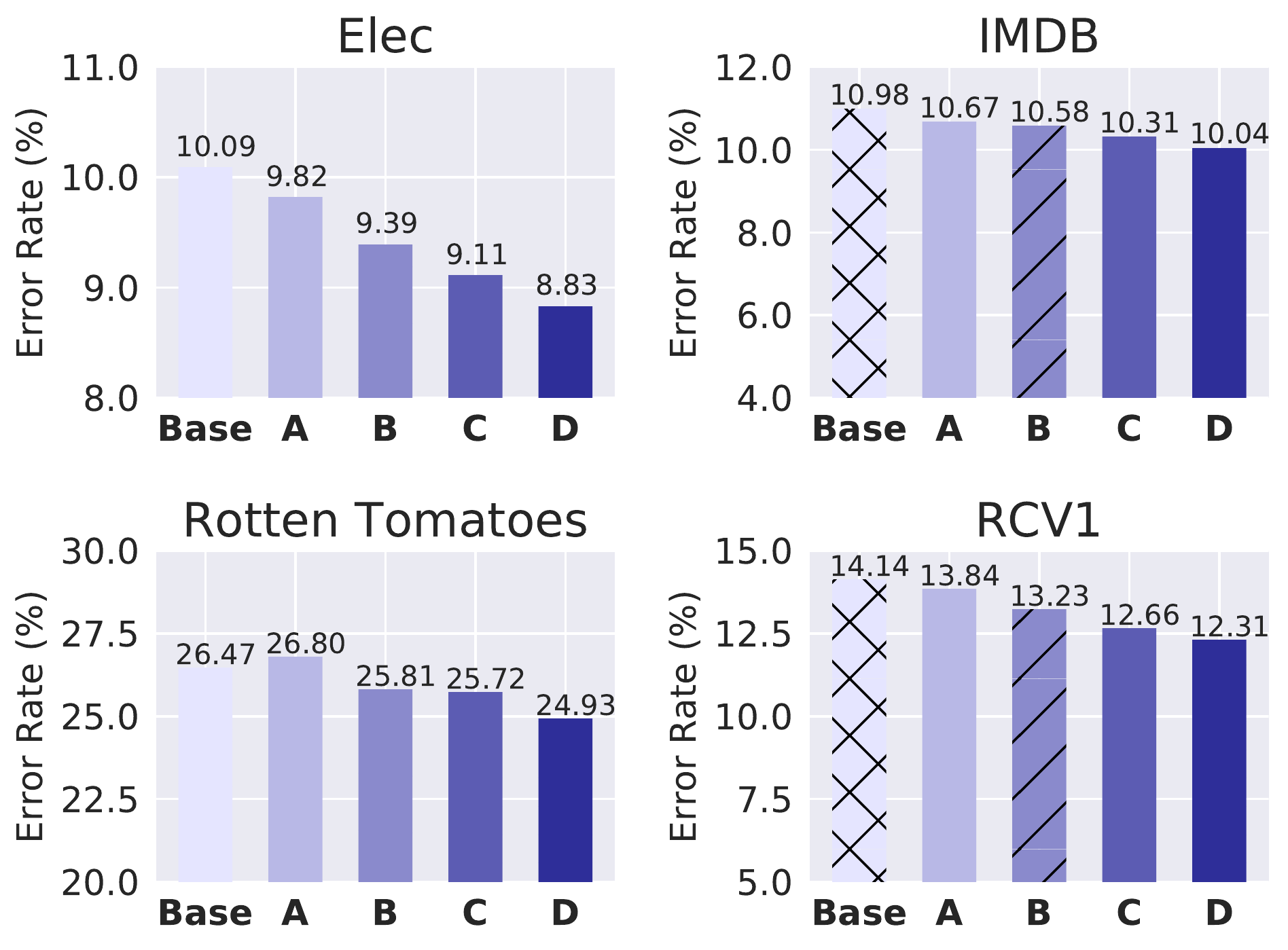}
    \caption{Effect of the \imitator{} with different window size $c_{i}$ on the final error rate (\%) of \weakbase{}. \textbf{A lower error rate indicates better performance.} \textbf{Base}: \expert{} (\weakbase{}) without the \imitator{}, \textbf{A}: $c_{i}=1$, \textbf{B}: $c_{i}=1,2$, \textbf{C}: $c_{i}=1,2,3$, \textbf{D}: $c_{i}=1,2,3,4$}
    \label{fig:numberweak}
\end{figure}

\begin{figure}[t!]
    \center
    \includegraphics[width=\hsize]{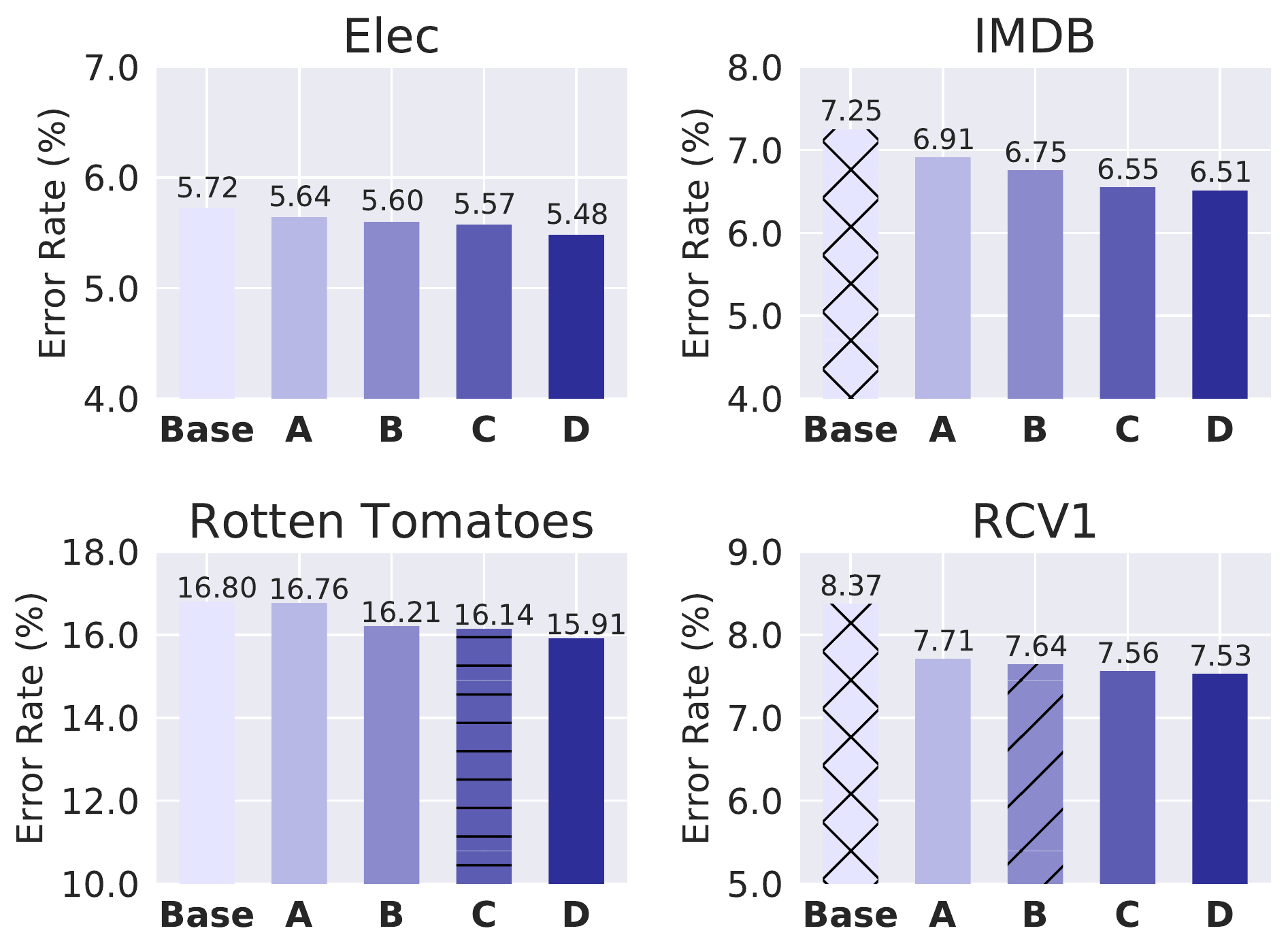}
    \caption{Effect of the \imitator{} with different window size $c_{i}$ on the final error rate (\%) of \strongbase{}. \textbf{A lower error rate indicates better performance.} \textbf{Base}: \expert{} (\strongbase{}) without the \imitator{}, \textbf{A}: $c_{i}=1$, \textbf{B}: $c_{i}=1,2$, \textbf{C}: $c_{i}=1,2,3$, \textbf{D}: $c_{i}=1,2,3,4$}
    \label{fig:numberstrong}
\end{figure}

Following Section~\ref{section:windowsize}, we investigated the effectiveness of combining the \imitator{}s with different window sizes ($c_{i}$) on the final error rate (\%) of the \expert{}.
We carried out experiment for both \weakbase{}+\imitator{} (Figure~\ref{fig:numberweak}) and \strongbase{}+\imitator{} (Figure~\ref{fig:numberstrong}).
The result is consistent to that of \adv{}+\imitator{} (Figure~\ref{fig:numberadv}), that greater window size improves the performance.

\end{document}